\documentclass{article}

\usepackage{microtype}
\usepackage{graphicx}
\usepackage{booktabs}
\usepackage{hyperref}
\usepackage[preprint]{icml2026}
\makeatletter
\renewcommand{\Notice@String}{\textit{Accepted to the 2nd Workshop on Compositional Learning at ICML 2026, Seoul, South Korea. Copyright 2026 by the author(s).}}
\makeatother
\hypersetup{pdfsubject={Accepted to the 2nd Workshop on Compositional Learning at ICML 2026, Seoul, South Korea}}

\usepackage[utf8]{inputenc}
\usepackage[T1]{fontenc}
\usepackage{url}
\usepackage{amsfonts}
\usepackage{amsmath}
\usepackage{amssymb}
\usepackage{nicefrac}
\usepackage{xcolor}
\usepackage{tikz}
\usepackage{pgfplots}
\pgfplotsset{compat=1.17}
\usepgfplotslibrary{groupplots}
\usetikzlibrary{positioning,arrows.meta,decorations.pathmorphing,calc,backgrounds}

\icmltitlerunning{When to Re-Plan: Subgoal Persistence in Hierarchical Latent Reasoning}

\begin{document}

\twocolumn[
  \icmltitle{When to Re-Plan: Subgoal Persistence in Hierarchical Latent Reasoning}

  \begin{icmlauthorlist}
    \icmlauthor{Ayushi Chadha}{ind}
  \end{icmlauthorlist}

  \icmlaffiliation{ind}{Independent Researcher, India}
  \icmlcorrespondingauthor{Ayushi Chadha}{ayushichadha48@gmail.com}

  \vskip 0.3in
]

\printAffiliationsAndNotice{}

\begin{abstract}
Long-horizon reasoning requires a system to commit to medium-horizon intent
without becoming rigid: re-plan too often and computation never coheres into
multi-step structure; commit too long and the plan goes stale. We study this
stability--adaptivity tradeoff in the latent reasoning setting, where multi-step
computation occurs inside hidden state rather than externalized token traces. We
extend the Hierarchical Reasoning Model (HRM) with a feudal-style manager--worker
interface: a slow high-level module periodically emits a normalized directional
subgoal that persists for $P$ low-level steps, biasing the worker's hidden-state
updates and supplying an intrinsic cosine alignment loss. On ARC and ConceptARC,
we find that subgoal \emph{persistence} --- not subgoal injection alone --- is
the central knob: moderate periods $P \in [3, 6]$ consistently outperform both
very frequent ($P{=}1$) and longer horizons, with a clear minimum LM loss at
$P{=}3$ ($1.544$ vs.\ $1.674$ at $P{=}1$ and a $1.640$ baseline; replicated
over five seeds with mean $1.595$ and standard deviation $0.045$). The intrinsic alignment weight $\lambda$ shows a complementary narrow optimum
($\lambda \approx 0.05$), and a controlled ablation at past-sweet-spot
$\lambda$ isolates \emph{learned} directional structure --- not architectural
capacity or auxiliary loss alone --- as the source of interference when the
alignment signal exceeds its optimum. Together
these findings implicate a design principle for compositional planning in latent
reasoning systems: medium-horizon intent must be coherent across enough
computational steps for compositional structure to form.
\end{abstract}

\section{Introduction}
\label{sec:intro}

A long-standing observation in cognitive science is that human reasoning
operates on at least two distinct timescales: a fast, automatic, intuition-driven
mode and a slow, deliberate, effortful mode~\citep{kahneman2011thinking,
evansstanovich2013dualprocess}. Modern reasoning systems in machine learning have
begun to recapitulate something like this structure. Chain-of-thought
prompting~\citep{wei2022cot} makes deliberation explicit by externalizing it as a
token sequence; latent reasoning architectures internalize multi-step computation
inside hidden state. Among the latter, adaptive computation
time~\citep{graves2016act} allocates variable depth as needed, and the
Hierarchical Reasoning Model
(HRM)~\citep{wang2025hrm} couples a slow high-level loop with a fast low-level
loop and a halting mechanism, achieving deep latent computation in compact models
trained from scratch. These architectures gain speed and flexibility by giving up
the legibility of the reasoning trace --- but in doing so, they raise a question
the externalized setting does not have to confront.

When deliberation is a sequence of explicit tokens, each token implicitly commits
the system to a piece of intent: the next word constrains what comes after. When
deliberation is a sequence of hidden-state updates, no such commitment is forced.
A latent reasoner can in principle revise its medium-horizon intent at every
micro-step, or never. The architectural form does not, by itself, determine the
temporal scale of planning. \textbf{How long should a latent reasoner commit to
an intent before revising it?}

We study this question through the lens of \emph{subgoal persistence}. One way
to give a latent reasoner explicit medium-horizon intent is to add a slow module
that periodically emits a subgoal --- a directional vector in latent space ---
that persists across multiple fast-module steps before being revised.
The choice of how long to commit to a subgoal is then the central knob, and it
embodies a stability--adaptivity tradeoff that any planner organized around
intermediate intent must resolve. Update too frequently and each subgoal is
overwritten before the worker can compose it into multi-step computation; commit
too long and the subgoal ossifies as the worker's hidden state evolves past it.
This tradeoff has been studied extensively in feudal reinforcement
learning~\citep{vezhnevets2017feudal,suttonprecupSingh1999options}, where it
underwrites the design of options and manager--worker hierarchies. We import this
commitment-duration lens into latent reasoning, where the worker's ``actions''
are hidden-state updates rather than environment interactions and the cost of
stale plans is representational rather than behavioral.

We propose \textbf{Subgoal-Augmented HRM}, a feudal-style extension of HRM that
operationalizes persistent latent intent. Every $P$ low-level micro-steps, the
high-level module emits a normalized directional subgoal $g$; the worker's
hidden-state updates are biased by $g$ through a learned projection, and a
cosine alignment loss rewards net low-level displacement that progresses along
$g$. Crucially, $P$ --- the manager period --- is exposed as a hyperparameter we
sweep, allowing us to characterize the persistence regime in which subgoals
actually help. We complement the persistence sweep with two further studies: a
sweep over the alignment loss weight $\lambda$ that distinguishes ``subgoal as
prior'' from ``subgoal as constraint,'' and a controlled ablation that isolates
the contribution of \emph{learned} directional structure from architectural
capacity alone.

Our findings on ARC and ConceptARC point to a single design principle.
Persistence is necessary, not optional: at $P{=}1$, the full subgoal
infrastructure performs \emph{below} the no-subgoal baseline --- the cleanest
evidence in our study that the persistence mechanism, not the injection
mechanism, drives the improvement. The benefit emerges sharply at $P{=}3$ and decays
only gradually out to $P{=}8$, an asymmetry suggesting that staleness is a
softer failure mode than instability. The intrinsic weight $\lambda$ shows a
complementary narrow optimum at $\lambda \approx 0.05$, consistent with the
mechanism functioning as a lightweight planning prior rather than a dominant
geometric constraint. A controlled ablation at past-sweet-spot $\lambda$
isolates \emph{learned} directional structure --- not architectural capacity or
auxiliary loss alone --- as the source of interference when the alignment
signal exceeds its optimum. Results are replicated across $5$ seeds at the best
configuration.

\paragraph{Scope.} This work is an empirical and behavioral characterization of
when and how persistent directional subgoals improve hierarchical latent
reasoning, complemented by a controlled ablation that isolates the source of
past-sweet-spot interference. A representation-level analysis of how persistent
subgoals shape worker hidden-state geometry is a natural complement we leave for
follow-up work; we discuss this in Section~\ref{sec:conclusion}.

We frame this as a contribution to compositional learning. The manager--worker
decomposition gives an explicit interface between intent and execution, allowing
the worker to assemble a sequence of subgoal-conditioned hidden-state updates
into longer-horizon computation. The persistence finding identifies the
\emph{necessary condition} under which this assembly actually occurs. A subgoal
mechanism without persistence has the architectural form of compositionality
(manager, worker, intent vector) but not its function: it cannot produce the
temporally extended structure that distinguishes ``planning a sequence of
steps'' from ``taking a single step with extra signal.'' The $P{=}1$ result is
the cleanest evidence of this gap.

\paragraph{Contributions.}
\begin{enumerate}
    \item We characterize a stability--adaptivity tradeoff in subgoal
    persistence for hierarchical latent reasoning, identifying $P \in [3, 6]$
    as a consistent sweet spot on ARC/ConceptARC and $P{=}1$ as underperforming
    the no-subgoal baseline. We replicate the best configuration across $5$
    seeds (Sections~\ref{sec:method}--\ref{sec:experiments}).
    \item We show that the alignment loss weight $\lambda$ exhibits a
    complementary narrow optimum at $\lambda \approx 0.05$, consistent with
    directional subgoals acting as a lightweight planning prior rather than a
    dominant constraint (Section~\ref{sec:experiments}).
    \item We provide a controlled ablation at past-sweet-spot $\lambda$ that
    isolates learned directional structure --- not architectural capacity or
    auxiliary loss alone --- as the source of interference when the alignment
    signal exceeds its optimum (Section~\ref{sec:ablation}).
\end{enumerate}

\section{Background and Related Work}
\label{sec:background}

\paragraph{Latent reasoning and adaptive computation.} Chain-of-thought
prompting~\citep{wei2022cot} externalizes reasoning as token sequences, gaining
interpretability and trainability at the cost of latency and brittleness to step
ordering. Latent reasoning architectures internalize multi-step computation in
hidden state. Adaptive Computation Time~\citep{graves2016act} allocates variable
depth as needed by learning a halting probability, increasing effective depth
when the input warrants it. The Hierarchical Reasoning Model
(HRM)~\citep{wang2025hrm} combines multi-timescale recurrence with halting:
coupled fast and slow loops exchange state through learned dynamics, with an
ACT-style halting head choosing the number of segments executed per example.
HRM achieves deep latent computation in compact models trained from scratch on
abstract reasoning tasks. However, coordination between the two levels in HRM is
implicit: there is no explicit signal carrying medium-horizon intent from the
slow loop to the fast one, and the temporal scale of any planning that occurs
must emerge from the recurrent dynamics alone.

\paragraph{Feudal hierarchies and temporal abstraction.} The gap left implicit in
HRM --- an explicit medium-horizon intent signal between hierarchical levels ---
is precisely the design problem that feudal hierarchies in reinforcement
learning address. The framework we extend originates in feudal reinforcement
learning, where a manager emits latent goals and a worker is incentivized to
follow them~\citep{vezhnevets2017feudal}. The
broader principle that temporally extended actions admit semi-Markov treatment
and provide credit-assignment leverage is formalized by the options
framework~\citep{suttonprecupSingh1999options}. Both lines of work establish that
\emph{commitment duration} is a first-class design choice: a manager that
re-issues goals every step is functionally not a manager, and an option without
a stable termination condition is not an option. We import this
commitment-duration lens from RL planning into latent reasoning, treating the
worker's hidden-state updates as the analogue of environment-interaction
``actions'' and the manager's emission period as the analogue of an option's
expected duration.

\paragraph{Compositional planning.} Subgoal decomposition is a canonical
mechanism by which planners achieve compositional generalization in feudal and
options-based reinforcement
learning~\citep{vezhnevets2017feudal,suttonprecupSingh1999options}: a complex
task is solved by composing solutions to simpler subgoals over extended
horizons. We adopt this lineage in latent reasoning, treating each persistent
subgoal as a directional primitive that the worker composes --- through a
sequence of subgoal-conditioned hidden-state updates --- into longer-horizon
computation. The stability--adaptivity tradeoff we characterize is, in this
view, a tradeoff about how long each primitive must remain in force for
composition to occur.

\section{Method}
\label{sec:method}

\subsection{HRM Backbone}
\label{sec:hrm-backbone}

We build on HRM with two coupled latent states: a slow high-level state $z^H_t$
and a fast low-level state $z^L_t$. Given an encoded input $\tilde{x}_t$, the
model performs iterative latent computation within a segment using recurrent
transformer blocks. At each micro-step $t$, the low-level state updates:
\begin{equation}
    z^L_{t+1} = f_L(z^L_t, z^H_t, \tilde{x}_t; \theta_L). \label{eq:lowlevel}
\end{equation}
The high-level state updates every $T$ low-level steps:
\begin{equation}
    z^H_{t+1} =
    \begin{cases}
        f_H(z^H_t, z^L_{t+1}; \theta_H), & (t+1) \bmod T = 0, \\
        z^H_t, & \text{otherwise}.
    \end{cases}
    \label{eq:highlevel}
\end{equation}
After $N$ micro-steps, the segment produces a prediction
$\hat{y} = f_O(z^H_N; \theta_O)$. Here $T$ is HRM's intrinsic high-level update
period --- a small fixed integer, typically $T{=}2$ in our experiments
following~\citet{wang2025hrm} --- and is independent of the manager period
$P$ we introduce in Section~\ref{sec:subgoal-emission} for subgoal emission.
HRM training uses deep supervision across segments with state detached between
segments, and an ACT-style halting objective~\citep{graves2016act} chooses the
number of executed segments per example. We treat the halting objective as
part of the HRM baseline.

\subsection{Subgoal Emission and the Persistence Period $P$}
\label{sec:subgoal-emission}

We add a feudal manager--worker interface inside HRM's latent computation. The
central design choice is the \textbf{manager period} $P$: the number of
low-level steps a subgoal persists before being revised. This parameter
controls the stability--adaptivity tradeoff directly --- small $P$ permits
rapid adaptation but prevents temporally extended computation from cohering
around any single intent; large $P$ provides commitment but risks staleness as
the worker's hidden state evolves past the issued goal.

Every $P$ low-level micro-steps, at times $t_k = kP$, the high-level state
emits a directional subgoal:
\begin{equation}
    \tilde{g}_k = W_g\, z^H_{t_k},
    \qquad
    g_k = \frac{\tilde{g}_k}{\|\tilde{g}_k\|_2 + \varepsilon}. \label{eq:subgoal-emit}
\end{equation}
A scalar commitment gate $\alpha_k = \sigma(w_\alpha^\top z^H_{t_k}) \in (0, 1)$
optionally modulates the strength of the subgoal's influence, allowing the
manager to soften commitment when the high-level state is uncertain. For steps
$t \in [t_k, t_k + P)$, the active goal and gate are $g(t) = g_k$ and
$\alpha(t) = \alpha_k$.

\paragraph{Why direction, not target.} A directional goal encodes ``where to
move'' in latent space without forcing the worker to hit a specific absolute
target --- a target-state formulation would be brittle under the nonstationary
internal dynamics of a recurrent reasoner. Direction is the minimal structure
that lets the manager commit to medium-horizon intent without overspecifying
execution.

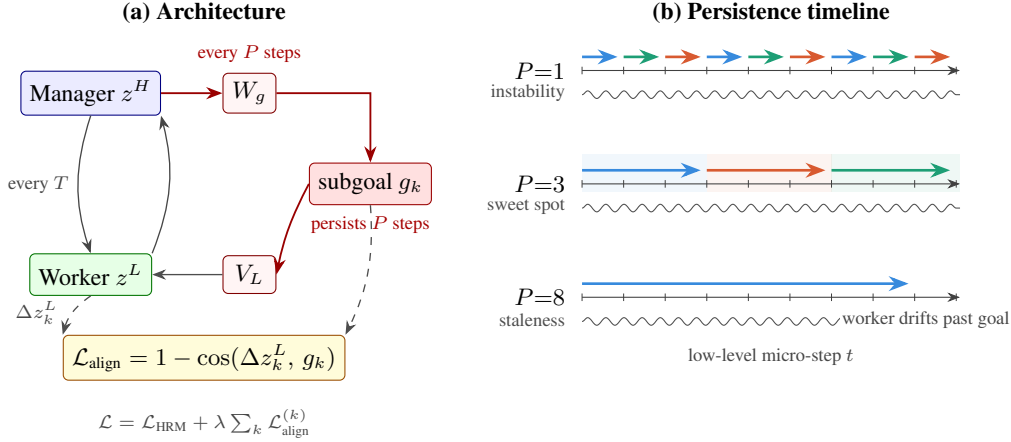
\begin{figure*}[t]
    \centering
    \begin{tikzpicture}[
        font=\footnotesize,
        block/.style={draw, rounded corners=2pt, minimum height=0.55cm,
                      align=center, fill=gray!5, line width=0.4pt},
        mgr/.style={block, fill=blue!8, draw=blue!50!black, minimum width=1.4cm},
        wkr/.style={block, fill=green!10, draw=green!45!black, minimum width=1.4cm},
        sub/.style={block, fill=red!12, draw=red!60!black, minimum width=1.2cm},
        proj/.style={block, fill=red!4, draw=red!50!black, minimum width=0.7cm,
                     minimum height=0.5cm},
        loss/.style={block, fill=yellow!18, draw=orange!60!black,
                     minimum width=2.0cm, minimum height=0.5cm},
        ar/.style={->, >=Stealth, line width=0.5pt},
        sig/.style={->, >=Stealth, line width=0.7pt, red!60!black},
        flow/.style={->, >=Stealth, line width=0.5pt, gray!60!black, dashed},
    ]

    \begin{scope}[local bounding box=arch]
      \node[mgr] (zh) at (0, 1.2) {Manager $z^H$};
      \node[wkr] (zl) at (0, -1.2) {Worker $z^L$};

      \draw[ar, gray!55!black]
          (zh.south) to[bend right=18]
          node[midway,left,font=\scriptsize,gray!55!black] {every $T$}
          (zl.north);
      \draw[ar, gray!55!black]
          (zl.north east) to[bend right=18] (zh.south east);

      \node[proj] (wg) at (2.1, 1.2) {$W_g$};
      \node[anchor=south,font=\scriptsize,red!65!black]
          at ($(wg.north)+(0,0.02)$) {every $P$ steps};

      \node[sub] (g) at (3.7, 0) {subgoal $g_k$};
      \node[anchor=north,font=\scriptsize,red!65!black]
          at ($(g.south)+(0,-0.02)$) {persists $P$ steps};

      \node[proj] (vl) at (2.1, -1.2) {$V_L$};

      \node[loss, minimum width=3.0cm] (lalign) at (1.5, -2.3)
          {$\mathcal{L}_{\text{align}} = 1 - \cos(\Delta z^L_k,\, g_k)$};

      \draw[sig] (zh.east) -- (wg.west);
      \draw[sig] (wg.east) -| (g.north);

      \draw[sig] (g.west) to[bend right=8] (vl.east);
      \draw[sig, gray!60!black, line width=0.5pt] (vl.west) -- (zl.east);

      \draw[flow] (zl.south) to[bend right=20]
          node[midway, left, font=\scriptsize, gray!55!black, xshift=-2pt] {$\Delta z^L_k$}
          (lalign.north west);
      \draw[flow] (g.south) to[bend left=15] (lalign.north east);

      \node[anchor=north, font=\scriptsize, gray!55!black]
          at (1.5, -2.85)
          {$\mathcal{L} = \mathcal{L}_{\text{HRM}} + \lambda \sum_k \mathcal{L}^{(k)}_{\text{align}}$};

      \node[anchor=south, font=\footnotesize\bfseries]
          at (1.5, 2.0) {(a) Architecture};
    \end{scope}

    \begin{scope}[xshift=6.5cm, local bounding box=tl]
      \def\xstart{0}
      \def\dx{0.55}
      \def\xend{5.0}

      \definecolor{sgA}{RGB}{55, 138, 221}
      \definecolor{sgB}{RGB}{216, 90, 48}
      \definecolor{sgC}{RGB}{29, 158, 117}

      \newcommand{\timelinerow}[3]{%
        \node[anchor=east, font=\footnotesize] at (-0.10, #1) {$#2$};
        \node[anchor=east, font=\scriptsize, gray!60!black]
            at (-0.10, #1-0.30) {#3};
        \draw[->, >=Stealth, gray!50!black, line width=0.4pt]
            (0, #1) -- (\xend, #1);
        \foreach \i in {0,...,9} {
            \draw[gray!50!black, line width=0.3pt]
                (\i*\dx, #1-0.05) -- (\i*\dx, #1+0.05);
        }
      }

      \newcommand{\subgoalarrow}[4]{%
        \draw[->, >=Stealth, line width=1.0pt, color=#3]
            (#1, #2+0.18) -- ($(#1, #2+0.18) + (#4*\dx-0.08, 0)$);
      }

      \newcommand{\workerline}[1]{%
        \draw[decorate,
              decoration={snake, amplitude=0.5mm, segment length=2.4mm},
              gray!60!black, line width=0.4pt]
            (0, #1-0.32) -- (\xend, #1-0.32);
      }

      \timelinerow{1.5}{P{=}1}{instability}
      \workerline{1.5}
      \subgoalarrow{0.00}{1.5}{sgA}{1}
      \subgoalarrow{0.55}{1.5}{sgC}{1}
      \subgoalarrow{1.10}{1.5}{sgB}{1}
      \subgoalarrow{1.65}{1.5}{sgA}{1}
      \subgoalarrow{2.20}{1.5}{sgC}{1}
      \subgoalarrow{2.75}{1.5}{sgB}{1}
      \subgoalarrow{3.30}{1.5}{sgA}{1}
      \subgoalarrow{3.85}{1.5}{sgC}{1}
      \subgoalarrow{4.40}{1.5}{sgB}{1}

      \timelinerow{0.0}{P{=}3}{sweet spot}
      \workerline{0.0}
      \subgoalarrow{0.00}{0.0}{sgA}{3}
      \subgoalarrow{1.65}{0.0}{sgB}{3}
      \subgoalarrow{3.30}{0.0}{sgC}{3}
      \begin{scope}[on background layer]
        \fill[sgA, opacity=0.07] (0.00, -0.10) rectangle (1.65, 0.40);
        \fill[sgB, opacity=0.07] (1.65, -0.10) rectangle (3.30, 0.40);
        \fill[sgC, opacity=0.07] (3.30, -0.10) rectangle (\xend, 0.40);
      \end{scope}

      \timelinerow{-1.5}{P{=}8}{staleness}
      \workerline{-1.5}
      \subgoalarrow{0.00}{-1.5}{sgA}{8}
      \node[anchor=south, font=\scriptsize, gray!55!black,
            fill=white, inner sep=1pt]
          at (4.55, -1.95) {worker drifts past goal};

      \node[anchor=north, font=\scriptsize, gray!55!black]
          at (\xend/2, -2.05) {low-level micro-step $t$};

      \node[anchor=south, font=\footnotesize\bfseries]
          at (\xend/2, 2.0) {(b) Persistence timeline};
    \end{scope}

    \end{tikzpicture}
    \caption{Subgoal-Augmented HRM. \textbf{(a)} Architecture. The manager
    (high-level state $z^H$, blue) projects through $W_g$ to emit a normalized
    directional subgoal $g_k$ every $P$ low-level steps. The subgoal is
    injected into the worker's update via $V_L$ (additive bias), persisting
    for $P$ steps before re-emission. Worker displacement $\Delta z^L_k$ over
    the commitment window is compared to $g_k$ via $\mathcal{L}_{\text{align}}$,
    added to the HRM objective with weight $\lambda$. Grey arrows indicate
    HRM's standard slow--fast recurrent coupling. \textbf{(b)} Persistence
    timeline over nine micro-steps. Coloured arrows above each axis denote
    subgoal emissions; their length spans the persistence window. At $P{=}1$
    a fresh direction is issued every step (no temporal coherence); at $P{=}3$
    three windows accumulate aligned worker displacement (shading); at $P{=}8$
    a single subgoal persists while the worker drifts past it. The wavy line
    sketches the worker's hidden-state trajectory $z^L$.}
    \label{fig:architecture}
\end{figure*}

\subsection{Subgoal Injection}
\label{sec:subgoal-injection}

The active goal is injected as an additive bias into the worker's update. Let
$V_L$ be a learned projection into the low-level latent space:
\begin{equation}
    z^L_{t+1} = f_L\!\left(z^L_t,\; z^H_t,\; \tilde{x}_t + \alpha(t)\, V_L\, g(t);\; \theta_L\right).
    \label{eq:lowlevel-inject}
\end{equation}
Optionally, the goal is also injected at the high-level update step via a
projection $V_H$:
\begin{equation}
    \begin{aligned}
    z^H_{t+1} &= f_H\!\left(z^H_t,\; z^L_{t+1} + \alpha(t)\, V_H\, g(t);\; \theta_H\right), \\
    &\hspace{3.2em} \text{when } (t+1) \bmod T = 0.
    \end{aligned}
    \label{eq:highlevel-inject}
\end{equation}
The injection converts medium-horizon intent into a persistent steering term
that shapes multiple consecutive worker updates, rather than relying on
emergent cross-level interactions to carry intent forward.

\subsection{Cosine Alignment Loss}
\label{sec:alignment-loss}

Persistent injection alone biases each update but does not guarantee the
worker's trajectory progresses \emph{along} the issued direction. We add a
cosine-based intrinsic loss --- the \emph{feudal alignment loss} in the
terminology of~\citet{vezhnevets2017feudal}, which we abbreviate to
\emph{alignment loss} for the remainder of this paper --- that rewards net
low-level displacement aligned with the active subgoal. Over the interval
$[t_k, t_k + P)$, define $\Delta z^L_k = z^L_{t_k + P} - z^L_{t_k}$, the net
displacement of the worker's hidden state during the subgoal's commitment
window. The alignment loss is:
\begin{equation}
    \mathcal{L}^{(k)}_{\text{align}} = 1 - \cos\!\left(\Delta z^L_k,\, g_k\right)
    = 1 - \frac{\langle \Delta z^L_k,\, g_k \rangle}{\|\Delta z^L_k\|_2\, \|g_k\|_2}.
    \label{eq:alignment}
\end{equation}
With optional gating, $\mathcal{L}^{(k)}_{\text{align}} = \alpha_k\bigl(1 - \cos(\Delta z^L_k, g_k)\bigr)$.

\subsection{Full Training Objective}
\label{sec:objective}

Let $\mathcal{L}_{\text{HRM}}$ denote the baseline HRM objective including task
loss and halting losses, unchanged from~\citet{wang2025hrm}. We add the
intrinsic alignment term, summed over the $\lfloor N/P \rfloor$ subgoal windows
within the current segment of $N$ micro-steps and weighted by $\lambda$:
\begin{equation}
    \mathcal{L} = \mathcal{L}_{\text{HRM}} + \lambda \sum_{k=1}^{\lfloor N/P \rfloor} \mathcal{L}^{(k)}_{\text{align}}.
    \label{eq:total-loss}
\end{equation}
Because HRM detaches state between segments, the alignment loss is computed
within each segment and does not propagate gradients across segment boundaries,
consistent with HRM's truncated-gradient training.

\section{Experiments}
\label{sec:experiments}

\subsection{Experimental Regimes}
\label{sec:regimes}

We report results from two complementary experimental setups, distinguished
by batch size, hardware, and dataset-augmentation factor. The \textbf{main
study} (Sections~\ref{sec:setup} through~\ref{sec:conceptarc-mini})
characterizes the mechanism's behavior across hyperparameters --- the
persistence sweep over $P$, the alignment-weight sweep over $\lambda$, and
the 5-seed replication of the best configuration --- using a global batch size of $768$ on CPU on the
\texttt{arc-aug-1000} corpus (\texttt{num\_aug=1000}). The \textbf{ablation
study} (Section~\ref{sec:ablation}) characterizes the causal contribution of
\emph{learned} directional structure under controlled conditions, using a
global batch size of $64$ on a single NVIDIA L4 GPU on a comparable corpus
constructed from the same source datasets but with reduced augmentation
(\texttt{num\_aug=100}) for tractability under the smaller batch size. Within
each setup, all hyperparameters other than the variables under study are
held fixed, and all reported numbers come from the same metric
($\texttt{train/lm\_loss}$) on identically-sized batches at matched
optimization steps.

The two setups produce different absolute LM-loss values for the same nominal
configuration --- for example, $\lambda{=}0.10, P{=}4$ yields $1.636$ in the
main study and $1.327$ in the ablation study --- reflecting differences in
batch-size optimization dynamics and effective dataset size rather than any
change in the model or objective. The \emph{magnitude} of past-sweet-spot
interference also differs: at $\lambda{=}0.10$ the main study shows the
mechanism barely matching baseline ($1.636$ vs.\ $1.640$), while the ablation
study at the same $\lambda$ shows the mechanism clearly underperforming
baseline ($1.327$ vs.\ $1.227$), consistent with smaller batches making
auxiliary-loss interference more visible. We therefore make all comparisons
\emph{within} their respective study and do not interpret absolute differences
\emph{across} studies as meaningful. The two studies are nonetheless
qualitatively consistent: both place $\lambda{=}0.10$ at or below the
no-subgoal baseline within their own setup, consistent with $\lambda{=}0.10$
lying past the optimum identified in the main $\lambda$ sweep
(Section~\ref{sec:lambda}). The ablation study extends this picture by
isolating \emph{why} the past-sweet-spot configuration interferes.

\subsection{Setup}
\label{sec:setup}

\paragraph{Data.} Our primary training corpus is derived from
ARC-AGI~\citep{chollet2019measure} and ConceptARC~\citep{moskvichev2023conceptarc},
using the \texttt{arc-aug-1000} configuration. Each puzzle is represented as a
flattened $30{\times}30$ grid sequence over a vocabulary of $12$ symbols
(padding, end-of-sequence, and ten colors). Each puzzle is augmented up to
$1000$ times during training via dihedral transformations (rotations and
reflections), random color permutations, and translational padding-based
shifts, following standard ARC augmentation practice. For cross-task
validation we additionally report results on ConceptARC-mini, a held-out
subset of $\sim$$4{,}900$ training and $\sim$$400$ validation samples grouped
by puzzle type.

\paragraph{Architecture.} HRM backbone with hidden size $512$, $4$ high-level
and $4$ low-level transformer layers, $8$ attention heads, recurrent cycles
$H_{\text{cycles}} = L_{\text{cycles}} = 2$, ACT with maximum $16$ internal
steps, RoPE positional encoding, SwiGLU feedforward layers, and RMS
normalization. The subgoal head projects $z^H$ to a $512$-dimensional
directional vector. All comparisons within a study use matched compute and
identical backbone configurations; the only variation is the subgoal
mechanism's hyperparameters ($P$ and $\lambda$).

\paragraph{Optimization.} We use the AdamATan2 optimizer with base learning
rate $10^{-4}$ and weight decay $0.1$, with puzzle-specific embeddings on a
separate learning rate of $10^{-2}$. A cosine schedule with $2000$ warmup
steps controls the learning rate.

\paragraph{Metrics.} We report LM loss as the primary metric: token-level
stablemax cross-entropy, masked over valid output positions, computed on the
training set at matched optimization steps. Secondary metrics include
token-level accuracy, exact accuracy (all output tokens correct), the alignment
loss itself, and average ACT halting depth.

\paragraph{Compute.} The main study reported in this section was conducted on
CPU with global batch size $768$. The ablation study reported in
Section~\ref{sec:ablation} was conducted on a single NVIDIA L4 GPU (24 GB)
hosted on E2E Networks with global batch size $64$. Section~\ref{sec:regimes}
discusses how this difference is handled across the paper.

\subsection{Persistence Is the Central Knob}
\label{sec:persistence}

\begin{figure*}[t]
    \centering
    \begin{tikzpicture}
    \begin{groupplot}[
        group style={
            group size=2 by 1,
            horizontal sep=1.2cm,
        },
        width=0.47\linewidth,
        height=4.5cm,
        tick label style={font=\footnotesize},
        label style={font=\footnotesize},
        title style={font=\footnotesize\bfseries, yshift=-2pt},
        every axis plot/.append style={line width=0.7pt},
        grid=both,
        grid style={gray!18},
        major grid style={gray!28},
    ]

    \nextgroupplot[
        title={(a) Persistence sweep ($\lambda{=}0.05$)},
        xlabel={Manager period $P$},
        ylabel={LM loss $\downarrow$},
        xmin=0.5, xmax=8.5,
        ymin=1.52, ymax=1.70,
        xtick={1,2,3,4,5,6,8},
        ytick={1.55,1.60,1.65,1.70},
    ]
    \fill[red!10, opacity=0.6]
        (axis cs:0.5,1.640) rectangle (axis cs:8.5,1.70);
    \fill[green!10, opacity=0.6]
        (axis cs:0.5,1.52) rectangle (axis cs:8.5,1.640);

    \addplot[gray!60!black, dashed, thick, domain=0.5:8.5, samples=2, forget plot] {1.640};
    \node[anchor=east, font=\scriptsize, gray!65!black]
        at (axis cs:7.8, 1.652) {baseline ($1.640$)};

    \addplot[blue!60!black, very thick, mark=*, mark size=2pt]
        coordinates {
            (1, 1.674) (2, 1.638) (3, 1.544) (4, 1.564)
            (5, 1.590) (6, 1.564) (8, 1.568)
        };

    \addplot[
        blue!60!black, only marks, mark=none, forget plot,
        error bars/.cd, y dir=both, y explicit,
        error bar style={line width=0.7pt}
    ] coordinates {
        (3, 1.544) +- (0, 0.045)
        (4, 1.564) +- (0, 0.048)
    };

    \addplot[red, only marks, mark=*, mark size=3pt,
             mark options={fill=red,draw=red}]
        coordinates {(3, 1.544)};

    \draw[->, >=Stealth, red!70!black, line width=0.5pt]
        (axis cs:1.7, 1.685) -- (axis cs:2.85, 1.560);
    \node[anchor=west, font=\scriptsize, red!70!black]
        at (axis cs:1.7, 1.692) {$\Delta {=} {-}0.13$};
    \node[anchor=south west, font=\scriptsize, red!70!black]
        at (axis cs:1.0, 1.674) {worst};
    \node[anchor=north west, font=\scriptsize, red!80!black]
        at (axis cs:3.1, 1.544) {best};

    \nextgroupplot[
        title={(b) Alignment-weight sweep ($P{=}4$)},
        xlabel={Alignment weight $\lambda$},
        ylabel={},
        xmode=log, log basis x=10,
        xmin=0.005, xmax=0.7,
        ymin=1.55, ymax=1.71,
        xtick={0.01,0.05,0.1,0.2,0.5},
        xticklabels={$0.01$,$0.05$,$0.1$,$0.2$,$0.5$},
        ytick={1.56,1.60,1.64,1.68},
        axis y line*=left,
    ]
    \addplot[gray!60!black, dashed, thick, domain=0.005:0.7, samples=2, forget plot] {1.640};
    \node[anchor=west, font=\scriptsize, gray!65!black]
        at (axis cs:0.006, 1.648) {baseline};

    \addplot[blue!60!black, very thick, mark=*, mark size=2pt]
        coordinates {
            (0.01, 1.612) (0.05, 1.569) (0.10, 1.636)
            (0.20, 1.662) (0.50, 1.691)
        };

    \addplot[red, only marks, mark=*, mark size=3pt,
             mark options={fill=red,draw=red}]
        coordinates {(0.05, 1.569)};
    \node[anchor=east, font=\scriptsize, red!80!black]
        at (axis cs:0.045, 1.569) {best};

    \draw[gray!55!black, line width=0.4pt, dotted]
        (axis cs:0.10, 1.555) -- (axis cs:0.10, 1.636);
    \node[anchor=north, font=\scriptsize, gray!55!black]
        at (axis cs:0.10, 1.560) {ablation $\lambda$};
    \end{groupplot}

    \begin{axis}[
        at={(group c2r1.south west)},
        anchor=south west,
        width=0.47\linewidth, height=4.5cm,
        xmode=log, log basis x=10,
        xmin=0.005, xmax=0.7,
        ymin=0.62, ymax=0.74,
        xtick=\empty,
        ytick={0.65,0.68,0.71},
        yticklabel style={font=\footnotesize},
        label style={font=\footnotesize},
        axis y line*=right,
        axis x line=none,
        hide x axis,
        ylabel={Accuracy $\uparrow$},
        xlabel={},
        every axis plot/.append style={line width=0.7pt},
        yticklabel pos=right,
    ]
    \addplot[orange!75!black, very thick, dashed, mark=square*, mark size=1.8pt,
             mark options={fill=orange!75!black, draw=orange!75!black}]
        coordinates {
            (0.01, 0.672) (0.05, 0.710) (0.10, 0.680)
            (0.20, 0.677) (0.50, 0.652)
        };
    \node[anchor=north east, font=\scriptsize, draw=gray!40, fill=white,
          rounded corners=1pt, inner sep=2pt, line width=0.3pt]
        at (axis cs:0.65, 0.738) {
          \begin{tabular}{@{}l@{\;}l@{}}
            \textcolor{blue!60!black}{\rule{0.4cm}{0.7pt}} & LM loss \\
            \textcolor{orange!75!black}{\rule{0.4cm}{0.7pt}} & Accuracy
          \end{tabular}
        };
    \end{axis}
    \end{tikzpicture}
    \caption{Main study results on ARC-AGI/ConceptARC. \textbf{(a)} Persistence
    sweep at $\lambda{=}0.05$. The shaded green/red regions mark below/above the
    no-subgoal baseline ($1.640$). $P{=}1$ underperforms baseline (worst); the
    benefit emerges sharply at $P{=}3$ ($\Delta{=}{-}0.13$, red point) and
    decays only gradually out to $P{=}8$ --- staleness is a softer failure
    mode than instability. Error bars at $P{=}3$ and $P{=}4$ are 5-seed
    standard deviations. \textbf{(b)} Alignment-weight sweep at $P{=}4$.
    LM loss (blue, left axis) shows a narrow optimum at $\lambda{=}0.05$,
    with token-level accuracy (orange dashed, right axis) tracking the same
    optimum. The ablation regime ($\lambda{=}0.10$) lies past the optimum and
    is the focus of Section~\ref{sec:ablation}.}
    \label{fig:persistence}
\end{figure*}
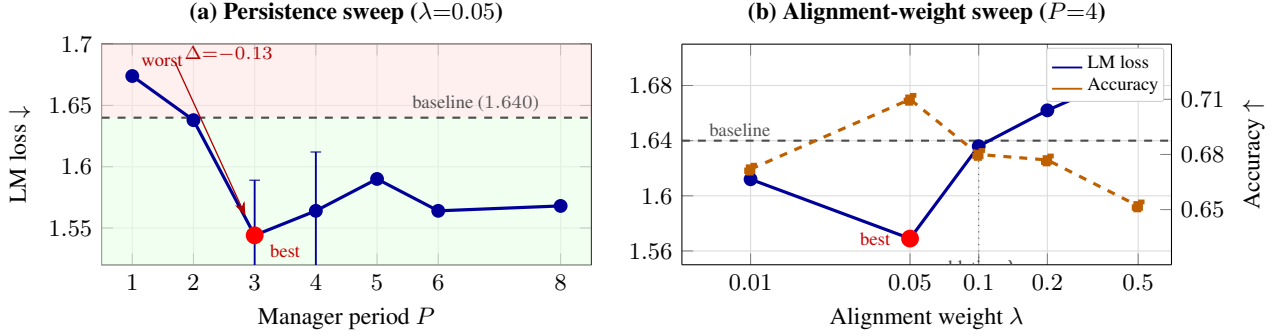

We sweep the manager period $P$ at fixed alignment-loss weight
$\lambda{=}0.05$, holding all other hyperparameters fixed.
Figure~\ref{fig:persistence}(a) plots the sweep curve.

Three observations are central to our claim.

\paragraph{Persistence is necessary, not optional.} At $P{=}1$ --- full subgoal
infrastructure but no commitment beyond a single step --- LM loss is $1.674$,
worse than the $1.640$ baseline without any subgoal mechanism. A subgoal that
is overwritten every step does not function as a subgoal. This is the cleanest
evidence in our study that the persistence mechanism, not the injection
mechanism, drives the improvement.

\paragraph{The benefit emerges sharply between $P{=}2$ and $P{=}3$.} $P{=}2$
improves over baseline only marginally ($1.638$ vs.\ $1.640$). The transition
to $P{=}3$ drops LM loss by $0.094$ --- by far the largest single step in the
sweep. This suggests a minimum coherence horizon below which subgoals cannot
organize compositional structure.

\paragraph{Decay beyond the optimum is gradual.} From $P{=}3$ to $P{=}8$, LM
loss varies in a narrow band $[1.544, 1.590]$, with no monotone degradation.
Subgoals tolerate moderate over-commitment but are substantially harmed by
under-commitment. We interpret this asymmetry as evidence that staleness is a
softer failure mode than instability: a slightly stale subgoal still provides
directional consistency, while an instantaneously revised subgoal provides
none.

\paragraph{Replication across seeds.} We replicate the two best configurations
across $5$ independent training runs each. At $\lambda{=}0.05, P{=}3$: mean LM
loss $1.595$ (std $0.045$, min $1.540$, max $1.662$). At $\lambda{=}0.05,
P{=}4$: mean LM loss $1.601$ (std $0.048$). The seed-level variability is
small relative to the gap between $P{=}3$ ($1.544$) and $P{=}1$ ($1.674$) and
between $P{=}3$ and the no-subgoal baseline ($1.640$), confirming that the
persistence finding is not a single-seed artifact.

\subsection{The Alignment Weight $\lambda$ Has a Narrow Optimum}
\label{sec:lambda}

Fixing $P{=}4$, we sweep the alignment-loss weight $\lambda$.
Figure~\ref{fig:persistence}(b) plots LM loss and token-level accuracy
against $\lambda$.

The optimum at $\lambda{=}0.05$ is narrow: $\lambda{=}0.10$ already approaches
the no-subgoal baseline, and $\lambda \geq 0.20$ degrades below it. We
interpret this as evidence that the alignment loss is most useful as a
\emph{prior} that gently shapes worker dynamics, rather than as a
\emph{constraint} that dominates task optimization. When the intrinsic
geometric signal competes with the task gradient on roughly equal terms
($\lambda \geq 0.10$), the worker's representational flexibility is reduced
enough to harm performance. This complements the persistence finding: subgoals
work best when they are committed-to long enough to organize computation
($P \in [3, 6]$) but soft enough not to override it ($\lambda \approx 0.05$).
We additionally verified that removing $\ell_2$ normalization of $g_k$ or the
commitment gate $\alpha_k$ degrades performance, confirming the directional
semantics and soft-prior role of the mechanism; a more substantive ablation
separating learned directional structure from the architectural addition
appears in Section~\ref{sec:ablation}.

\subsection{Cross-Task Validation on ConceptARC-mini}
\label{sec:conceptarc-mini}

To check whether the persistence effect generalizes beyond the
\texttt{arc-aug-1000} corpus, we evaluate on ConceptARC-mini, a held-out task
family grouped by puzzle type. The no-subgoal baseline achieves LM loss
$2.316$; Subgoal-Augmented HRM at $\lambda{=}0.05, P{=}3$ achieves $2.308$.
The improvement is modest in absolute terms ($\sim$$0.4\%$), and we therefore
treat this as a directionally consistent secondary observation rather than
independent benchmark evidence: the headline persistence and $\lambda$ effects
characterized above remain the load-bearing claims.

\section{Past-Sweet-Spot Ablation: What Causes Interference?}
\label{sec:ablation}

The main $\lambda$ sweep (Section~\ref{sec:lambda}) characterizes the
\emph{symptom} of past-sweet-spot interference but does not identify its
\emph{source}. We address this with a controlled three-cell ablation at
$\lambda{=}0.10$ --- past the optimum, where interference is reliably
observable. The cells (Table~\ref{tab:ablation-flags}) differ only in three
flags governing the subgoal mechanism, isolating learned directional
structure as the variable of interest while controlling for architectural
capacity and the auxiliary loss term.

\begin{table}[t]
    \centering
    \caption{Ablation conditions at $\lambda{=}0.10$, $P{=}4$. All three
    cells share an identical HRM backbone, optimizer, dataset, and training
    duration, and differ only in the three subgoal-mechanism flags shown.
    \textbf{A\_full} is Subgoal-Augmented HRM as proposed.
    \textbf{B\_baseline} disables both injection and the alignment loss,
    reducing the model to vanilla HRM with no manager--worker interface.
    \textbf{E\_random} keeps both active but replaces the manager's
    \emph{learned} directions with random unit vectors, isolating the
    contribution of learned directional content from the architectural
    addition and auxiliary loss term.}
    \label{tab:ablation-flags}
    \resizebox{\columnwidth}{!}{%
    \begin{tabular}{lccc}
        \toprule
        Cell & \texttt{inject\_subgoal} & \texttt{use\_alignment\_loss} & \texttt{random\_directions} \\
        \midrule
        A\_full       & true  & true  & false \\
        B\_baseline   & false & false & false \\
        E\_random     & true  & true  & true  \\
        \bottomrule
    \end{tabular}%
    }
\end{table}

\subsection{Setup}
\label{sec:ablation-setup}

\paragraph{Hardware, batch size, and data.} Training was performed on a
single NVIDIA L4 GPU with global batch size $64$, in contrast to the main
study's CPU runs at batch size $768$ (Section~\ref{sec:regimes}). All three
cells trained on the same ARC-AGI plus ConceptARC augmented corpus
(\texttt{num\_aug=100}, \texttt{seed=42}; $95{,}993$ puzzle identifiers),
constructed from the same source corpora as the main study but with reduced
augmentation factor for tractability under the smaller batch size.

\paragraph{Hyperparameters and training duration.} All three cells used
$\lambda{=}0.10$, $P{=}4$, learning rate $10^{-4}$, $500$ training epochs,
and a single fixed random seed. The HRM backbone and optimizer match
Section~\ref{sec:setup}. All three cells terminated at the same optimization
step ($30{,}321$ of an intended $31{,}143$, the last full epoch boundary
that fit within the iteration budget) under identical software conditions ---
this shared termination is a property of the training driver and makes the
three cells directly comparable. We report training-set LM loss (the same
metric as the main study), under the within-regime comparison rule described in
Section~\ref{sec:regimes}.

\subsection{Results}
\label{sec:ablation-results}

\begin{table}[t]
    \centering
    \caption{Train LM loss at step $30{,}321$ for the three ablation cells.
    The within-regime contrast structure --- not the absolute values --- is
    what the result speaks to.}
    \label{tab:ablation-results}
    \begin{tabular}{lcc}
        \toprule
        Cell & Train LM loss $\downarrow$ & $\Delta$ vs.\ B\_baseline \\
        \midrule
        A\_full       & $1.327$ & $+0.100$ \\
        B\_baseline   & $1.227$ & --- \\
        E\_random     & $1.230$ & $+0.003$ \\
        \bottomrule
    \end{tabular}
\end{table}

Three within-regime contrasts are central.

\paragraph{A vs.\ B: at $\lambda{=}0.10$, the full mechanism interferes.} Adding
learned directional subgoals raises LM loss by $\sim$$0.10$ relative to the
no-subgoal baseline, consistent with the main $\lambda$ sweep
(Section~\ref{sec:lambda}, Figure~\ref{fig:persistence}(b)) where $\lambda{=}0.10$
already performs at or below baseline.

\paragraph{E vs.\ B: random directions are nearly indistinguishable from
baseline.} When learned directions are replaced with random unit vectors,
LM loss matches baseline within $0.003$ --- an order of magnitude smaller
than the seed-level variability ($0.045$) observed in the main study's
replication. The architectural addition and auxiliary loss term are not, by
themselves, harmful at $\lambda{=}0.10$.

\paragraph{A vs.\ E: interference is specifically driven by \emph{learned}
directional content.} The $0.097$ gap between A\_full and E\_random ---
much larger than the $0.003$ gap between E\_random and baseline --- isolates
the interference as specifically due to the \emph{learned} directional
content of the subgoals, not the additional capacity, auxiliary objective,
or injection mechanism per se.

\subsection{Interpretation}
\label{sec:ablation-interpretation}

This ablation refines the picture established by the $\lambda$ sweep. The
narrow optimum at $\lambda{\approx}0.05$ shows \emph{that} the alignment loss
functions best as a soft prior; the ablation isolates \emph{why} it stops
functioning as a soft prior past the optimum. The issue is not generic
auxiliary-loss interference --- random directions, with the same loss term,
do not interfere. Rather, learned directional structure is doing real
representational work in A\_full (which is why it helps at $\lambda{=}0.05$
and hurts at $\lambda{=}0.10$); when the alignment weight competes with the
task gradient on roughly equal terms, the worker is pulled toward directions
that serve alignment but not the LM objective. This contrast identifies learned directional structure as the active
component: it helps near the optimum but becomes harmful when over-weighted.

\section{Discussion}
\label{sec:discussion}

The persistence sweep identifies a regime in which medium-horizon intent
measurably helps a latent reasoner, and identifies the failure modes outside
it. We organize the discussion around the stability--adaptivity tradeoff.

\paragraph{The persistence tradeoff.} A subgoal revised every step provides
no temporal structure for the worker to compose: each micro-update receives
a fresh, potentially uncorrelated direction, the alignment loss provides
high-variance gradients with no medium-horizon signal, and the injection
adds noise the worker must compensate for. This is why $P{=}1$ underperforms
baseline. Between $P{=}3$ and $P{=}6$, subgoals persist long enough for
multiple consecutive updates to accumulate aligned displacement, and the
alignment loss rewards consistent progress rather than instantaneous
alignment. The asymmetry between under- and over-commitment ($P{=}1$
catastrophic, $P{=}8$ only mildly degraded) suggests staleness is a soft
failure mode while absence of commitment is not: compositional structure
tolerates approximation in commitment duration but does not tolerate its
absence.

\paragraph{Why $\lambda$ has a narrow optimum, and what the ablation adds.}
The alignment loss is a structural prior, not a task signal. Too small, it
fails to shape latent dynamics; too large, it competes with the task gradient
and reduces representational flexibility. The narrow optimum at
$\lambda \approx 0.05$ is consistent with auxiliary objectives being most
effective as soft regularizers. The past-sweet-spot ablation
(Section~\ref{sec:ablation}) refines this: the conflict past the optimum is
not generic auxiliary-loss interference but specifically the conflict
introduced by \emph{learned} directional structure that, when over-weighted,
captures representational capacity the worker would otherwise use for the
task.

\section{Limitations and Conclusion}
\label{sec:conclusion}

Gains are modest in absolute terms, with evaluation restricted to ARC-style
tasks (held-out check on ConceptARC-mini); broader validation is needed.
The past-sweet-spot ablation uses a single fixed seed (train-set LM loss at
one step), though main-study replication variance ($\text{std}\approx 0.045$)
is much larger than the E vs.\ B gap, bounding single-seed variability well
below the contrast magnitude. A representation-level analysis of
subgoal-induced hidden-state geometry would directly probe the compositional
substrate but requires late-training checkpoints we did not retain; we leave
this for follow-up work. Subgoal \emph{persistence}, controlled by $P$, is
the design knob determining whether directional latent subgoals function in
HRM; the narrow $\lambda$ optimum and past-sweet-spot ablation together
identify the stability--adaptivity tradeoff any compositional planner in
latent space must resolve.

\section*{Impact Statement}
This paper presents work whose goal is to advance the field of Machine
Learning, specifically the design of compact latent reasoning systems with
explicit temporal abstraction. The experiments are conducted on abstract
reasoning benchmarks and do not involve human-subjects data, private data, or
deployed decision systems. We do not identify immediate negative societal
consequences beyond those generally associated with improved machine-learning
reasoning systems.

\bibliography{references}

@book{kahneman2011thinking,
  title     = {Thinking, Fast and Slow},
  author    = {Kahneman, Daniel},
  year      = {2011},
  publisher = {Farrar, Straus and Giroux},
  address   = {New York}
}

@article{evansstanovich2013dualprocess,
  title   = {Dual-Process Theories of Higher Cognition: Advancing the Debate},
  author  = {Evans, Jonathan St. B. T. and Stanovich, Keith E.},
  journal = {Perspectives on Psychological Science},
  volume  = {8},
  number  = {3},
  pages   = {223--241},
  year    = {2013}
}

@inproceedings{wei2022cot,
  title     = {Chain-of-Thought Prompting Elicits Reasoning in Large Language Models},
  author    = {Wei, Jason and Wang, Xuezhi and Schuurmans, Dale and Bosma, Maarten and Ichter, Brian and Xia, Fei and Chi, Ed H. and Le, Quoc V. and Zhou, Denny},
  booktitle = {Advances in Neural Information Processing Systems},
  volume    = {35},
  pages     = {24824--24837},
  year      = {2022}
}

@article{wang2025hrm,
  title   = {Hierarchical Reasoning Model},
  author  = {Wang, Guan and Li, Jin and Sun, Yuhao and Chen, Xing and Liu, Changling and Wu, Yue and Lu, Meng and Song, Sen and Yadkori, Yasin Abbasi},
  journal = {arXiv preprint arXiv:2506.21734},
  year    = {2025}
}

@article{graves2016act,
  title   = {Adaptive Computation Time for Recurrent Neural Networks},
  author  = {Graves, Alex},
  journal = {arXiv preprint arXiv:1603.08983},
  year    = {2016}
}

@inproceedings{vezhnevets2017feudal,
  title     = {{FeUdal} Networks for Hierarchical Reinforcement Learning},
  author    = {Vezhnevets, Alexander Sasha and Osindero, Simon and Schaul, Tom and Heess, Nicolas and Jaderberg, Max and Silver, David and Kavukcuoglu, Koray},
  booktitle = {Proceedings of the 34th International Conference on Machine Learning},
  series    = {Proceedings of Machine Learning Research},
  volume    = {70},
  pages     = {3540--3549},
  publisher = {PMLR},
  year      = {2017}
}

@article{suttonprecupSingh1999options,
  title   = {Between {MDP}s and Semi-{MDP}s: A Framework for Temporal Abstraction in Reinforcement Learning},
  author  = {Sutton, Richard S. and Precup, Doina and Singh, Satinder},
  journal = {Artificial Intelligence},
  volume  = {112},
  number  = {1--2},
  pages   = {181--211},
  year    = {1999}
}

@article{moskvichev2023conceptarc,
  title   = {The {ConceptARC} Benchmark: Evaluating Understanding and Generalization in the {ARC} Domain},
  author  = {Moskvichev, Arseny and Odouard, Victor Vikram and Mitchell, Melanie},
  journal = {Transactions on Machine Learning Research},
  year    = {2023}
}

@article{chollet2019measure,
  title   = {On the Measure of Intelligence},
  author  = {Chollet, Fran{\c{c}}ois},
  journal = {arXiv preprint arXiv:1911.01547},
  year    = {2019}
}
\bibliographystyle{icml2026}



\end{document}